# Comparison of SMT and RBMT, The Requirement of Hybridization for Marathi – Hindi MT

**Sreelekha .S, Pushpak Bhattacharyya**
Dept. of Computer Science and Engineering,
IIT Bombay
sreelekha@cse.iitb.ac.in     pb@cse.iitb.ac.in

## Abstract

We present in this paper our work on comparison between Statistical Machine Translation (SMT) and Rule-based machine translation for translation from Marathi to Hindi. Rule Based systems although robust take lots of time to build. On the other hand statistical machine translation systems are easier to create, maintain and improve upon. We describe the development of a basic Marathi-Hindi SMT system and evaluate its performance. Through a detailed error analysis, we, point out the relative strengths and weaknesses of both systems. Effectively, we shall see that even with a small amount of training corpus a statistical machine translation system has many advantages for high quality domain specific machine translation over that of a rule-based counterpart.

## 1   Introduction

Machine Translation is the process of translating text or speech from one natural language to another with the help of machines. There are many ongoing attempts to develop MT systems for various regional languages using rule-based as well as statistical-based approaches. MT systems can be called as a bilingual system if it is designed specifically for two particular languages and can be called as a multilingual system if it is designed for more than a single pair of languages. Development of efficient machine translation (MT) systems using appropriate methodologies and with limited resources is a challenging task. Since India is rich in linguistic divergence there are many morphologically rich languages quite different from English as well as from each other, there is a great need for machine translation between them (Nair et.al, 2010). It has 18 constitutional languages, which are written in 10 different scripts. Even though MT in India started more than two decades ago, it is still an ongoing process (Antony, 2013). This paper discusses various approaches used in Indian language to Indian language MT systems especially in Marathi – Hindi MT systems.

Handling the structural difference between the two languages and handling the ambiguity are the two major difficulties in Machine Translation.

### 1.1   Challenge of Ambiguity

There are two types of ambiguity: structural ambiguity and lexical ambiguity.

### 1.1.1 Lexical Ambiguity

Words and phrases in one language often have multiple meaning in another language.
  For example, in the sentence,

*Marathi-* मी फोटो काढला *{me photo kadhla}*
*Hindi-* मैने फोटो निकाला *{maenne photo nikala}*
*English-* ***I took the photo***

  Here in the above sentence "*काढला*"{kadhla}, "*निकाला*"*{nikala},* and "*took*" have ambiguity in meaning. It is not clear that whether the word "*काढला*"{kadhla} is used as the "clicked the photo" ("*निकाला*" *{'nikala'}* in Hindi) sense or the "took" (*nikala*) sense.
   However this is a good example where both in source language and target language ambiguity is present for the same word.

   This will usually be clear from the context, but this kind of disambiguation is generally non-trivial.

### 1.1.1 Structural Ambiguity

Due to the structural order there will be multiple meanings. For example,

Marathi - तिथे उंच मुली आणी मुलें होती .
{tithe oonch muli aani mulen hoti}
*{There were tall girls and boys there}*

Here from the words *"उंच मुली आणी मुलें "* { oonch muli aani mulen } it is clear that tall girls but it is not clear that boys are tall, since in Marathi to represent tall girls and boys only one word "उंच" {oonch} {tall} is being used. It can have two interpretations in Hindi and English according to it's structure.

Hindi - वहाँ लंबी लड़कियां और लडकें थे ।
*{vahan lambi ladkiyam our ladkem the }*

*{There were tall girls and boys there}*
or
Hindi - वहाँ लंबी लड़कियां और लंबे लडकें थे ।
*{vahan lambi ladkiyam our lambe ladkem the}*

*{There were tall girls and tall boys there}*

Handling this kind of structural ambiguity is one of the big problems in Machine Translation.

### 1.2 Structural Differences

In the case of Marathi – Hindi machine translation both languages follow the same structural ordering in sentences, such as Subject- Object-Verb (SOV). Even though there is ordering similarity, there are morphological and stylistic differences which have to be considered during translation.

Marathi is morphologically more complex than Hindi, wherein there are a lot of post-modifiers in the former as compared to the later (Dabre *et al*, 2012, Bhosale, 2011).

For example, the word form "सातपुरींच्या"*{saatpurimchya} {of / about the seven pilgrimage spots}* is derived by attaching "च्या"*{chyaa}* as a suffix to the plural form "सातपुरीं" *{saatpurim}* which is derived from the noun "सातपुरी" *{saatpuri}{seven pilgrimage spots in India}* by undergoing an inflectional process. Marathi exhibits agglutination of suffixes which is not present in Hindi and therefore these suffixes have equivalents in the form of post positions. For the above example, the Hindi equivalent of the suffix "च्या" *{chyaa}* is the post position "के"*{ke}* which is separated from the plural noun "सप्तपुरियों"*{saptapuriyom}*. Hence the translation of "सातपुरींच्या"*{saatpurimchya}* will be "सप्तपुरियों के" *{saptapuriyom ke}*.

Similarly the Marathi verb form "जाणार्या" *{janaaryaa} {the one who is going}* which is derived by affixing "णा"*{naa}* and "र्या" *{ryaa}* to the stem "जा"*{jaa} {go}* has a Hindi translation "जाने वाला"*{jaane wala}*.

In the case of sentence ordering both Marathi and Hindi follows SOV. For example,

Hindi- हरिद्वार को गंगा द्वार भी कहते हैं ।
{*haridwar ko ganga dwar bhi kehte hai* }
　(S)　　　(O)　　　(V)
Marathi- *हरिद्वारला गंगाद्वार असेही म्हटले जाते* .
　(S)　　　(O)　　　(V)
*English-{Haridwarala gangadwar asehi mehtale jaate}*

{*Haridwar is also known as Ganga* }
　(S)　　　(V)　　　(O)

This is an advantage for statistical machine translation system as we shall see a later section.

### 1.3 Vocabulary Differences

Languages differ in the way they lexically divide the conceptual space and sometimes no direct equivalents can be found for a particular word or phrase of one language in another.
Consider the sentence,
" काल आनंदीचे केळवण होते . "

*Here* " केळवण " as a verb has no equivalent in Hindi, and this sentence has to be translated as,
"कल आनंदी का सगाई होने के बाद एवं शादि के पहले लडका या लडकी को संबंधीयों द्वारा दिया जाने वाला भोज था ।"
*{"Kal aanandii ka sagaayi hone ke baad evam shaadi ke pahle ladka ya ladki ko sambandhiyon dwara diya jaane wala bhoj tha . " }*

{The lunch which the relatives are giving before marriage to bride or groom has been given to Ananthi yesterday}

tions, wherein Section 2 deals with Rule Based Machine Translation (RBMT), Section 3 deals with Statistical Machine Translation (SMT), Sec-

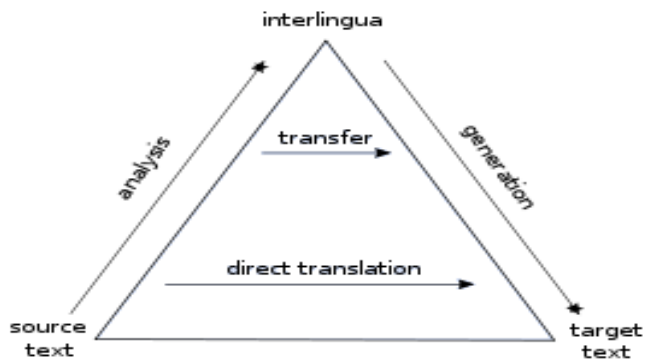

Figure 1: Vacquois Triangle

The obvious difficulty is in determining the translation of such language specific concepts which pose additional challenges in machine translation.

### 1.4 Different types of Machine Translation

The Vacquois triangle in the figure1 depicts three different types of Machine Translation namely, Transfer based, Interlingua based and Statistical. They differ in the amount of linguistic processing performed before transferring concepts and structure from the source side to the target side. As can be seen Interlingua requires complete processing, Transfer based requires some and Statistical (a type of direct translation) requires none. The base of the triangle indicates the distance between the two languages and linguistic processing helps bridge the gap.

The rest of the paper is divided into three sections, wherein Section 2 deals with Rule Based Machine Translation (RBMT), Section 3 deals with Statistical Machine Translation (SMT), Section 4 deals with Experiments conducted, Evaluations and Error analysis which concludes the main components of the paper.

## 2 Rule Based Machine Translation (RBMT)

Rule based MT systems works based upon specification of rules for morphology, syntax, lexical selection and transfer and generation. Collection of rules and a bilingual or multilingual lexicon are the resources used in RBMT. In the case of English to Indian languages and Indian language to Indian language MT systems, there have been many attempts with all these approaches (Dave et al., 2002). The transfer model involves three stages: analysis, transfer and generation. Each of them is described below. Refer to figure 2 for the complete flow of translation in the form of a pipeline.

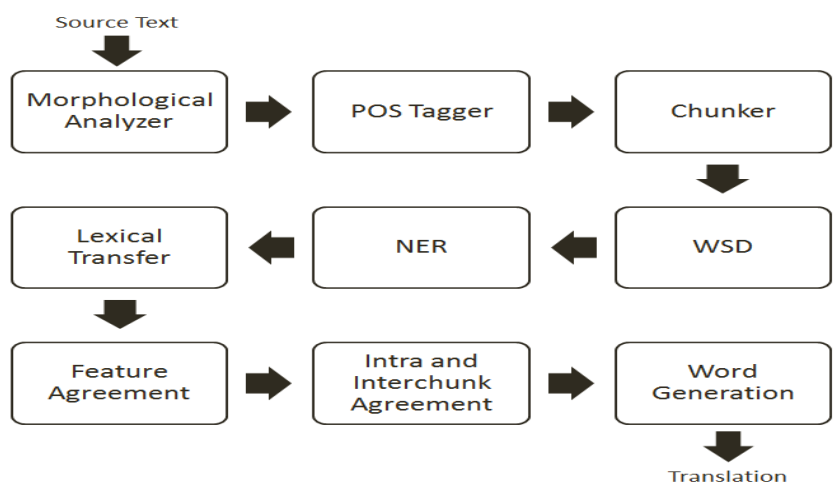

Figure 2: Rule based System

### 2.3.1 Analysis

During this phase, from the input text information about the morphology, parts of speech, shallow phrases, entity and word sense disambiguation information is extracted.

### 2.3.2 Lexical transfer

The lexical transfer phase involves two parts namely word translation and grammar translation which is performed using high quality bilingual dictionary and transfer grammar rules.

### 2.3.3 Generation phase

Generation involves correction of the genders of the translated words since certain words are masculine in the source language but feminine in the target and vice versa. This is followed by short distance and long-distance agreements performed by intra-chunk and the inter-chunk modules concluded by word generation.

## 3 Statistical Machine Translation (SMT)

The statistical approach comes under Corpus Based Machine Translation systems, which tries to generate translations based on the knowledge and statistical models extracted from parallel aligned bilingual text corpora. Statistical models take the assumption that every word in the target language is a translation of the source language words with some probability (Brown et al., 1993). The words which have the highest probability will give the best translation. Consistent patterns of divergence between the languages (Dorr et al., 1994, Dave et al., 2002, Ramananthan et al. , 2011) when translating from one language to another, handling reordering divergence are one of the fundamental problems in MT. Figure 3 shows the functional flow diagram of an SMT system. The three major steps in SMT are:

### 3.1 Corpus preparation

To prepare a properly cleaned and well aligned parallel corpus is the major requirement of an SMT system. The quality of the resultant translation will be based upon the quality of the parallel sentence translation in the source corpus. Corpus preparation, alignment and its cleaning is done in the Pre-Processing step.

### 3.2 Training

Training is a process in which a supervised or unsupervised statistical machine learning algorithms are used to build statistical tables from the parallel corpora (Zhang *et al*., 2006). In Statistical Machine Translation, word by word and phrase based alignment plays the major role during parallel corpus training. During training Translational model, Language Model, Distortion Table, Phrase table etcetera are modeled.

### 3.3 Decoding

Decoding (Och and Ney, 2001, Och and Ney, 2003, Koehn, 2007) is the most complex task in Machine Translation (Knight, 1999) where the trained models will be decoded. This is the main step in which the target language translations are being decoded using the generated phrase table and language model. Decoding complexity and target language reordering (Kunchukuttan and Bhattacharyya 2012) are the two major concerns with SMT.

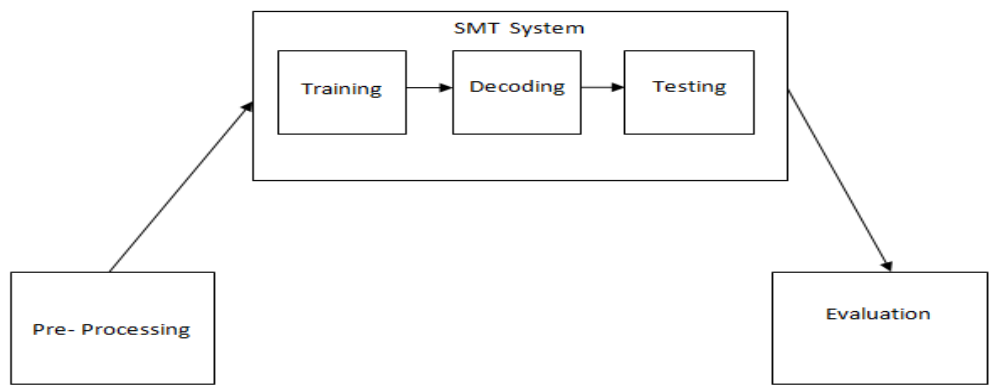

**Figure 3: SMT System**

# 4 Experimental Discussion

## 4.1 Statistical Machine Translation System Experiments

We now describe the development of our Marathi- Hindi SMT system, the experiments performed and the comparisons with the results, in the form of an error analysis, of the Rule Based system described above. For the purpose of constructing with statistical models we use Moses and Giza++[1].

### 4.1.1 Pre- processing

Usually parallel corpus available for Indian languages is not of sufficiently high quality which is quite important in order to improve the quality of the translation. The factors that indicate a poor quality corpus are six fold namely:

1. Misalignment between parallel sentences which prevents learning of word to word alignments.
2. Too many stylistic constructions in the parallel corpus which prevent the learning of grammatical structures in terms of phrases for small corpus.
3. Missing translations in corpus on source or target side.
4. Missing words and phrases in the corpus leading to incorrect alignments.
5. Unwanted characters present in corpus which leads to garbage in the output due to incorrect alignments.
6. Wrong translations in the corpus leading to wrong outputs.

We manually cleaned a 90000 sentence parallel corpus for Marathi- Hindi, corrected the grammatical structure of the sentences and tokenized it thereby making available a high quality corpus for training. We also injected 0.4 million Marathi- Hindi bilingual dictionary words extracted from the IndoWordnet [2] to the phrase table through the training corpus to increase the number of words that the SMT system can translate. Since Moses tokenization tools are not customized for Devanagari scripts, we tokenized the corpus using our own tokenization tools.

### 4.1.2 Training

Table 1 and 2 describe the complete resources we have used for training. The dictionary words were used to improve the quality of the translation system. It increased both the language model and phrase table size to a great extent. We followed the training steps of Moses baseline system.

| Sl.No | Corpus Source | Training Corpus [Manually cleaned and aligned] | Corpus Size [Sentences] |
|---|---|---|---|
| 1 | ILCI | Tourism | 25000 |
| 2 | ILCI | Health | 25000 |
| 3 | DIT | Tourism | 20000 |
| 4 | DIT | Health | 20000 |
| | | Total | 90000 |

Table 1: Statistics of Training Corpus

| Sl.No | Dictionary Source | Dictionary Words in Corpus | Dictionary Size [Words] |
|---|---|---|---|
| 1 | CFILT, IIT Bombay | Indo Wordnet Synset words | 400000 |
| | | Total | 400000 |

Table 2: Statistics of the Dictionary words

| Sl. No | Corpus Source | Testing corpus [Manually cleaned and aligned] | Corpus Size [Sentences] |
|---|---|---|---|
| 1 | EILMT | Tourism | 100 |
| | | Total | 100 |

Table 3: Statistics of Testing Corpus

---

[1] http://www.statmt.org/

[2] http://www.cfilt.iitb.ac.in/indowordnet/

**4.1.3 Testing**

We have tested the translation system with a corpus of 100 sentences taken from the 'EILMT tourism health' corpus as shown in Table 3. The added advantage was the SOV ordering similarity between Marathi and Hindi. However there were difficulties in handling inflected words. Since Hindi has comparatively less amount of inflections such as suffix words etcetera as compared to Marathi.

100 sentences from EILMT corpus for verifying the translation quality.

Consider the same sentence as before:

तो घरी जाणाऱ्यांबरोबर जाई.

{to ghari jaanaryaambarobar jaie}
{He used to go with the ones who used to go home}.

The Rule Based output for this sentence is

वह घर में कटने वाले के साथ जाता था।

{vah ghar maen katnevale ke saath jat tha }

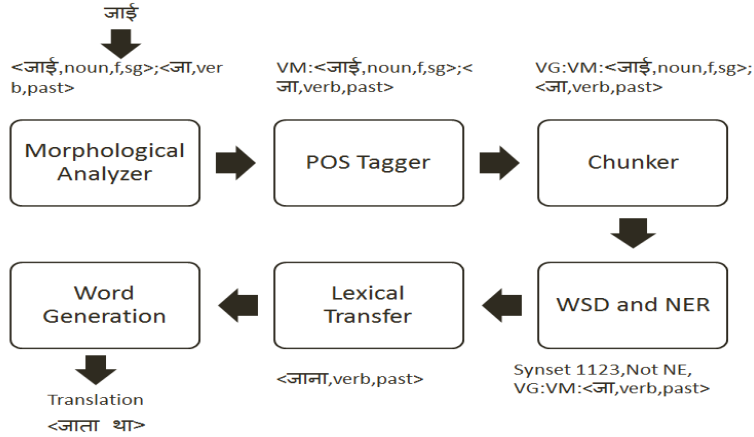

Figure 4: Work flow of RBMT

For example,

तो घरी जाणाऱ्यांबरोबर जाई.

{to ghari jaanaryaambarobar jaie}
{He used to go with the ones who used to go home}.

The SMT system is translated into Hindi as

वह घर जाणाऱ्यांबरोबर|UNK|UNK|UNK जाई.|UNK|UNK|UNK

{vah ghar jaanaryaambarobar jaie }
{He used to go with the ones who used to go home}

SMT system is not able to map the words "जाई" into "जाता था", "जाणाऱ्यांबरोबर|" into "जाने वालों के साथ" because these word forms were not present in the parallel corpus. Moreover enumerating all possibilities of inflected word forms is not possible manually. And thus the SMT system is faces difficulty in handling morphology.

**4.2 Rule-Based Machine Translation System Experiments**

Using the Marathi-Hindi Rule Based Machine Translation system described before we tested

Each word will be processed through the RBMT pipe line as shown like in figure 2. The important steps of the RB system flow for the word "जाई"{jaaie}{used to go} is given in the figure 4.

**1. Analysis**: The morphological analyzer identifies the word "जाई" {jaaie} as either a noun or a verb in past tense. After POS tagging, it is identified that the word is a Verb and the Chunker determines that it is a part of a Verb Group. After WSD the appropriate sense is determined.

**2. Transfer:** The lexical transfer module translates it to "जाना" {jaana} {to go}.

**3. Generation:** Since the sentence is short the agreement phenomenon is not so significant. The word generator takes the information about "past tense" to give the final word form: "जाता था" {jaata tha} {used to go}.

However the translation is far from good, considering that the translation of जाणाऱ्यांबरोबर {jaanaryaanbarobar} is कटने वाले के साथ

{katne wale ke saath} which is not accurate. Here the system is not able to accurately determine the correct translation sense of "जा"{ja} leading to a poor lexical choice. Also the plural information is lost as the suffix "वाले"{vale} is generated instead of "वालों"{valon}. In the output sentence the post position में {maen} {in} is not a fluent translation; its absence is preferred.

We observed that although, rule based MT was able to handle rich morphology, leading to meaning transfer, it was unable to effectively handle the appropriate translation and generation of function words and common word senses which are handled well by SMT, which improve fluency (Ahsan, et al. , 2010).

As can be seen from the above described example, the translation of a single word requires a number of steps, each involving considerable linguistic inputs. Hence we can come to a conclusion that the rule-based machine translation process is extremely time consuming, difficult, and fails to analyze accurately and quickly a large corpus of unrestricted text due to inherent errors in the modules which are part of the system.

### 4.3 Evaluation

In order to evaluate the quality of the translations we have used subjective evaluation to determine fluency (F) and adequacy (A). We did consider BLEU scores (Papineni et al.) for evaluation made and its grammatical correctness. The basis of scoring is given below:

- 5: If the translations are perfect.
- 4: If there are one or two incorrect translations and mistakes.
- 3: If the translations are of average quality, barely making sense.
- 2: If the sentence is barely translated.
- 1: If the sentence is not translated or the translation is gibberish.

S1, S2, S3, S4 and S5 are the counts of the number of sentences with scores from 1 to 5 and N is the total number of sentences evaluated. The formula (Bhosale et al., 2011) used for computing the scores is:

$$A/F = 100 * \frac{(S5 + 0.8 * S4 + 0.6 * S3)}{N}$$

We consider only the sentences with scores above 3. Moreover we penalize the sentences with scores 4 and 3 by multiplying their count by 0.8 and 0.6 respectively so that the estimate of scores is much better. As these scores are subjective, they vary from person to person in which case an inter annotator agreement is required. Since we had only one evaluator we do not give these scores. The results of our evaluations are given in Table 4 and Table 5.

| MT System | Adequacy | Fluency |
|---|---|---|
| Rule Based | 69.6% | 58% |
| Statistical | 62.8% | 73.4% |

Table 4: Results of Subjective Evaluation

| MT System | BLEU Score |
|---|---|
| Rule Based | 5.9 |
| Statistical | 9.31 |

Table 5: Results of BLEU score Evaluation

uation, however based on our studies it was deemed as an unreliable measure due to the morphological richness of both languages. Fluency is an indicator of correct grammatical constructions present in the translated sentence whereas adequacy is an indicator of the amount of meaning being carried over from the source to the target. For each translation we assigned scores between 1 and 5 depending on how much sense the trans-

### 4.4 Error Analysis

We have evaluated the translated outputs of both Rule Based and Statistical Machine Translation system. The detailed error analysis is shown in Table 6 for five sentences exhibiting a variety of linguistic phenomena.

| | Sr. No. | Sentence | Explanation of phenomena |
|---|---|---|---|
| 1 | Source Sentence | केंद्रीय सरकारी संग्रहालय १८७६मध्ये प्रिन्स औफ वेल्सच्या भारतभेटीच्या वेळी उभारण्यात आले व १८८६ साली ते जनतेसाठी खुले करण्यात आले. | In the rule based system since each word was morphologically analysed the overall meaning is conveyed however "1886 सालें" {1886 saale} {year (plural) 1886} is not a grammatically good construction. This is overcome in the SMT system by replacing it by a more fluent form "1886 में" {1886 mein}. Moreover the proper from of वह {waha} {it} is picked in the SMT system but not in the rule based system namely "वे" {wey} {they}. However, the content words are not translated in the SMT system due to lack of learned word forms. |
| | Meaning | In 1986 the national central museum was established during the visit of the Prince of Wales and in 1886 was opened for the public. | |
| | Rule based system | केंद्रीय सरकारी संग्रहालय 1876 में प्रिन्स औफ वेल्स के भारतभेट का बार में उठाया गया व 1886 सालें वे जनता के लिए खुला किया गया । | |
| | Statistical System | केंद्रीय सरकारी संग्रहालय १८७६मध्ये प्रिंस औफ वेल्सच्या भारतभेटीच्या के शेड डाला गया व १८८६ में वह जनता के लिए खोल दिया गया । | |
| 2 | Source Sentence | दीग पॅलेस भक्कम व प्रचंड किल्ला आहे, जो भरतपूरच्या शासकांचे ग्रीष्मकालीन निवासस्थान होता. | The RB system makes a mistake in sense disambiguation of the word "प्रचंड"{prachand}{huge} which also has the sense of many, which the SMT system does not. SMT is also able to overcome the number agreement between "का" and "ग्रीष्मकालीन" leading to a more fluent translation. Due to the morphological richness of Marathi "भरतपूरच्या" is translated correctly as "भरतपूर के" by RB system but not by SMT system (it gives "भरतपूरच्या के"). |
| | Meaning | Deeg palace, which was the rural era residence of the rulers of Bharatpur, is tough and huge. | |
| | Rule based system | दीग पॅलेस मजबूत व बहुत किला है , जो भरतपूर के शासकों के ग्रीष्मकालीन आवास हो । | |
| | Statistical System | दीग पैलेस मजबूत व विशाल किला है , जो भरतपूरच्या के शासकों का ग्रीष्मकालीन निवास था । | |
| 3 | Source Sentence | मारवाड हा राजस्थानमधील मुख्य उत्सव, ऑक्टोबर महिन्यामध्ये संप्पन्न होतो. | Since "मारवाड" was not present in the training corpus and the input dictionary the SMT system made a wrong translation. However function word translation of "मधील" {madhil} {of} is better done by the SMT system. Overall the RB translation is clear but not as fluent as the SMT system. |
| | Meaning | Marwad, a major festival in Rajasthan, takes place in the month of October. | |
| | Rule based system | मारवाड हा राजस्थान में के मुख्य उत्सव ऑक्टोबर महीने में संप्पन्न हो । | |
| | Statistical System | राजस्थान का यह राजस्थान का प्रमुख त्यौहार अक्टूबर के महीने में संप्पन्न होता है । | |
| 4 | Source Sentence | शेकडो आणि हजारो पर्यटक सॅम सँड ड्युनमध्ये निसर्गाचे मोहक कालात्मक दृश्य पहाण्याकरता राजस्थानला येतात व ते ठिकाण सर्वोत्तमरीत्या उंटावरील सफारीने पाहिले जाऊ शकते. | The sentence is quite long with many grammar constructions which both, the RB system and the SMT system are not able to handle well. However the translations of content words fares better in the RB translation. In this case many source words are not translated in the SMT output which adversely affects its |
| | Meaning | Hundreds of thousands of travellers come to Som Sand Dunes in Rajasthan to view the artistic sights and that place can be viewed best by a camel safari. | |
| | Rule based system | सैकड़ों और हजारों पर्यटक सॅम सँड ड्युन में प्रकृति की मोहक कालात्मक दृश्य पहाण या करता राजस्थान को आते हैं व वे स्थान सर्वोत्तमरी में या ऊँट पर के सफारी ने देखा जा सकता है । | fluency as well. This is one type of input which fares poorly when its SMT translation is compared to that of a RB system |
| | Statistical System | सैकड़ों और हजारों पर्यटक सॅम सँड ड्युनमध्ये प्रकृति का मोहक कालात्मक दृश्य पहाण्याकरता राजस्थान आते हैं और वह स्थान सर्वोत्तमरीत्या उंटावरील सफारीने देखा जा सकता है । | |
| 5 | Source Sentence | उदयपुरला अनेक मोहक आकर्षणे आणि नैसर्गिक सौंदर्याचे वरदान आहे, राजस्थानच्या प्रसिद्ध शहरांपैकी एक शहर, त्याच्या उत्कृष्ट शिल्पशास्त्र आणि हस्तकलेसाठी ओळखले जाते. | Here the SMT system makes quite a grave mistake by inserting additional wrong words like "गाँवों , कसबों" {gaavon, kasbon} {villages} as translations in place of "शहर" {shahar} {city}. Nevertheless overall the fluency, attributed by the proper handling of gender agreements of "का" {ka} {of} with "सौंदर्य" {saundarya} {beauty}. Naturally the overall quality seems less due to the lack of translation of the content words by the SMT system which is done better by the RB system. |
| | Meaning | Udaypur, is blessed with many tempting attractions and natural beauty, a renowned city of Rajasthan, is known for its excellent crafts and handicrafts. | |
| | Rule based system | उदयपुर को अनेक मोहक आकर्षण और प्राकृतिक सुंदरता की वर है , राजस्थान के प्रसिद्ध शहरों में से एक शहर , उसके उत्कृष्ट स्थापत्य और हस्तकले के लिए पहचाना जाता है। | |
| | Statistical System | उदयपुरला कई मोहक आकर्षण एवं प्राकृतिक सौंदर्य का वरदान है , राजस्थान के प्रसिद्ध शहरों में से एक गाँवों , कसबों , उसके उत्कृष्ट स्थापत्य और हस्तकलेसाठी जाना जाता है । | |

Table 6 : Error Analysis of Translation

We observed that translation quality of Statistical Machine Translation is relatively high as compared to the Rule Based system, considering that the efforts required to build RBMT systems is huge. The result of BLEU score evaluation is displayed in Table 5 and the result of Subjective evaluation is displayed in Table 4. The fluency of the SMT outputs was very good compared to RBMT indicated by a fluency of 73.4% for SMT system where as 58% for RBMT. The reason that the SMT system had a very high fluency was due to plentiful evidences of good quality phrase pairs recorded in the phrase table. Moreover the language model used, helped in generating more natural translations. But in terms of adequacy RBMT showed slight improvement as compared to SMT, since Marathi is a morphologically complex language. Also SMT which cannot split suffixes by itself was unable to handle the translation of suffix words in some cases. RBMT being able to use the morph analyzer, can easily separate the suffixes from the inflected words and generate translations inflected with correct gender number person, tense, aspect and mood (GNPTAM). However due to poor quality Word Sense Disambiguation incorrect translations are generated. This is mitigated by SMT since it records phrase translations with respect to frequency which acts as a more natural sense disambiguation mechanism.

## 5 Conclusion

In this paper we have mainly focused on the comparative performance of Statistical Machine Translation and Rule- Based Machine

Translation for Marathi - Hindi. As discussed in the experimental section, SMT, although lacks the ability to handle rich morphology, does not fall much behind RBMT. It has a staggering advantage over RBMT in terms of fluency and the ability to capture natural Hindi structure. This leads to the requirement of a hybridized approach for Machine Translation between Marathi and Hindi.

Our future work will be focused on the integration of Rule- Based system components namely the Morphological Analyses into the Statistical Machine Translation system and there by develop a Hybridized MT system for Marathi- Hindi Machine Translation.